\documentclass{article} 
\usepackage[final]{neurips_mlsb_2025}
\usepackage{times}

\usepackage{amsmath,amsfonts,bm}

\def\eqref#1{equation~\ref{#1}}

\def\1{\bm{1}}

\DeclareMathAlphabet{\mathsfit}{\encodingdefault}{\sfdefault}{m}{sl}
\SetMathAlphabet{\mathsfit}{bold}{\encodingdefault}{\sfdefault}{bx}{n}

\usepackage{epigraph}
\usepackage{bbm}

\usepackage{hyperref}
\usepackage{url}
\usepackage{mathtools}
\usepackage{algorithm}
\usepackage{algorithmic}
\usepackage{multirow}
\usepackage{multicol}
\usepackage{graphicx}
\usepackage{siunitx}
\usepackage{booktabs}
\usepackage{xcolor}
\usepackage{tabularx}
\usepackage{amsmath}
\usepackage{comment}
\usepackage{wrapfig}
\usepackage[nameinlink]{cleveref}

\title{On fine-tuning Boltz-2 for protein-protein affinity prediction}

\author{James King, Lewis Cornwall, Andrei Cristian Nica, James Day, Aaron Sim,\\ \textbf{Neil Dalchau, Lilly Wollman, Joshua Meyers}\\
    Synteny \\
    \textit{London, UK}
}

\begin{document}
\maketitle
\begin{abstract}

Accurate prediction of protein–protein binding affinity is vital for understanding molecular interactions and designing therapeutics. We adapt Boltz-2, a state-of-the-art structure-based protein–ligand affinity predictor, for protein–protein affinity regression and evaluate it on two datasets, TCR3d and PPB-affinity. Despite high structural accuracy, Boltz-2-PPI underperforms relative to sequence-based alternatives in both small- and larger-scale data regimes. Combining embeddings from Boltz-2-PPI with sequence-based embeddings yields complementary improvements, particularly for weaker sequence models, suggesting different signals are learned by sequence- and structure-based models. Our results echo known biases associated with training with structural data and suggest that current structure-based representations are not primed for performant affinity prediction.

\end{abstract}

\section{Introduction}

Protein–protein interactions (PPIs) underpin nearly all cellular processes, and accurate prediction of their binding affinities is essential for understanding molecular mechanisms and guiding therapeutic design~\citep{lu2020recent}. Computational methods for affinity prediction are increasingly critical, offering scalable alternatives to experimental approaches~\citep{ppi_review}. This is especially true in the development of immunomodulatory bispecifics, precise modelling of T cell receptor (TCR)–peptide–MHC and antibody–antigen interactions is particularly important, as small changes in affinity can determine clinical efficacy and safety~\citep{middelburg2021overcoming}.

Recent advances in machine learning have transformed multimeric protein structure prediction, exemplified by AlphaFold-Multimer~\citep{evans2021protein} and its successor AlphaFold 3 (AF3)~\citep{abramson2024accurate}. Open-source implementations such as Boltz~\citep{wohlwend2025boltz} have broadened accessibility, enabling large-scale exploration of protein–protein complex structures. Although work remains to predict challenging interfaces such as those mediating antibody or TCR target recognition, these tools represent substantial progress and lay the foundation for affinity prediction and downstream therapeutic applications.

\section{Background and Related Work}

Previous methods for predicting protein-protein affinity have taken advantage of both sequence-based features and structure-based representations~\citep{ppi_review}. 

Standardised benchmark datasets such as PPB-affinity~\citep{PPB_origonal} and the more recent PPB-affinity (filtered)~\citep{PPB_enhanced} provide an invaluable resource for systematically training and evaluating such methods. Related developments in protein–ligand affinity prediction, such as Boltz-2~\citep{passaro2025boltz}, demonstrate the potential of modern machine learning architectures, motivating their adaptation and extension to the protein–protein affinity setting.

Structure-based affinity prediction models, however, have been shown to exploit dataset biases by memorising entities or pocket geometry~\citep{Li2024} while sequence-based models risk learning global sequence similarity rather than binding determinants~\citep{Tsishyn2024}. In both cases, careful mitigation strategies must be employed to ensure models capture interactions, not artifacts.

In this work, we make minimal modifications to Boltz-2 to permit training of a protein-protein affinity model. We show using two datasets that training an structure-based affinity model is subject to pitfalls introduced by aspects of the data and contrast this with sequence-based alternatives. We discuss the hurdles that must be leapt in order to train a useful structure-based affinity predictor.

\section{Datasets}

\subsection{TCR3d}

TCR3d 2.0~\citep{TCR3d_reference} comprises TCR-pMHC complex structures curated from the Protein Data Bank (PDB). A subset of 251 of these are both complexed with class-I or class-II MHC molecules, and associated with affinity measurements (dissociation constant, $K_d$) obtained via surface plasmon resonance (SPR) or other biophysical techniques. We transform $K_d$ values to $pK_d$ and exclude three entries for which the TCR binding mode is peptide-agnostic (9eji, 9ejg and 9ejh).\footnote{as of October 1, 2025}

We design hard splits between our train, validation, and test splits using the algorithm presented in Appendix~\ref{app:dataset-split}. In this way, similar sequences are discouraged from appearing in different splits.

\subsection{PPB-affinity}

The PPB-affinity (filtered) dataset~\citep{PPB_enhanced} describes 8,207 unique protein-protein interaction entries with binding affinity measurements ($K_d$). This dataset is a filtered version of the original PPB-affinity dataset~\citep{PPB_origonal} which collates varied PPI data from multiple sources. Each entry is associated with a PubMed ID (PMID) indicating the source of the affinity experiment. The filtered dataset resolves annotation inconsistencies by removing duplicate entries from multi-chain protein interactions. The \textit{PPB-affinity} dataset contains many different types of proteins and therefore allows a more general model to be trained, considering protein-protein affinities beyond TCR-pMHC complexes.

We reuse splits described by the authors~\citep{PPB_enhanced}, which were generated by implementing a $\leq 30\%$ sequence identity threshold to split proteins into training, validation, and test sets. We reuse baseline results from this paper directly, these are sequence-based approaches which operate directly on protein sequences of the ligand and receptor to learn an association with affinity.

\section{Methods}
\subsection{Adapting Boltz-2 for protein-protein affinity prediction}

Boltz-2 extends Boltz-1~\citep{wohlwend2024boltz} by improving structure prediction and, crucially for this work, adding an affinity module trained to predict protein-ligand affinity. The single and pairwise Pairformer representations are passed through an \textit{affinity module}, which consists of a further distance-conditioned Pairformer stack, followed by cross-pair pooling and MLP readouts for affinity prediction.

The Boltz-2 affinity module is trained on protein-ligand interactions. In order to adapt the affinity module for protein-protein interactions, we make the following modifications.
\begin{enumerate}
    \item \textbf{Loss function.} We define the optimisation objective to be
        \begin{equation*}
        L(y, \hat{y}) \coloneqq \lambda L_{\text{Huber}}(y, \hat{y}) + (1-\lambda) L_{\text{rank}}^{\text{PMID}}(y, \hat{y}),
        \end{equation*}
        where $y$ and $\hat y$ denote a batch of true and predicted affinities, respectively. $L_{\text{Huber}}$ is the Huber loss,

        \begin{equation*}
        L_{\text{Huber}}(y, \hat{y}) \coloneqq \frac{1}{n}\sum_{i=1}^n \begin{cases}
        \frac{1}{2}(y_i - \hat{y}_i)^2 & \text{if } |y_i - \hat{y}_i| \leq \delta, \\
        \delta |y_i - \hat{y}_i| - \frac{1}{2}\delta^2 & \text{otherwise},
        \end{cases}
        \end{equation*}
        
        and $L_{\text{rank}}^{\text{PMID}}$ is a ranking loss between pairs within a batch 
        \begin{equation*}
        L_{\text{rank}}^{\text{PMID}}(y, \hat{y}) \coloneqq \frac{1}{n} \sum_{i=1}^n \sum_{j=1}^n \mathbbm{1}_{\hat y_i > \hat y_j}\log\left(1 + e^{y_i - y_j}\right).
        \end{equation*}

    \item \textbf{Batch construction.} The ranking loss function involves comparison of predicted affinities within a batch. In order to mitigate study-wise batch effects, we construct each batch from a single PMID only.
    \item \textbf{Affinity module input.} The Boltz-2 affinity module uses representations from both the inter-chain peptide-ligand interactions, and the intra-chain ligand-ligand interactions, but not intra-chain protein-protein interactions. In order to respect the symmetry of protein-protein interactions, we included \textit{all} inter- and intra-chain representations.
\end{enumerate}

We freeze the weights of the Boltz-2 trunk and trained the affinity module from scratch using He initialisation~\citep{he2015delving}.

\subsection{Combining sequence and structure representations}

We perform a simple concatenation of the MLP representations extracted from our Boltz-2-PPI affinity module with the final linear embeddings from the sequence models in \citep{PPB_enhanced} and construct a simple linear projection as our combined sequence-structure model. We selected the ESM2-650M \citep{lin2022language} and ProtT5 \citep{elnaggar2021prottrans} models as the two different architectures for our sequence models in our experiments.

\section{Results}

\subsection{Performance on the TCR3d Dataset}

We report the results of fine-tuning Boltz-2 on the TCR3d dataset in Table~\ref{tab:tcr3d-results}. Our fine-tuned model \textit{Boltz-2-PPI}, is shown alongside our sequence-based ESM2 650M baseline~\citep{lin2022language}, which was fine-tuned with an MSE loss on the same data. Unfortunately the sequence-based models achieve stronger predictive performance than Boltz-2-PPI.

A clear limitation is the small size of the dataset, which hinders effective training of a structure-based model. To mitigate this, we experimented with reducing the size of the affinity module to a simple 2-layer MLP, but this did not yield improvements.

We also investigated whether the quality of predicted structures was influencing the training of affinity models. We observed that Boltz-2 achieves high structure prediction accuracy (DockQ $\approx 0.91$) on TCR-pMHC structures that were present in its training set, but reduced accuracy (DockQ $\approx 0.70$) for unseen complexes~\citep{wohlwend2025boltz}. Since our affinity dataset spans both included and more recent complexes, we trained an additional model using experimentally determined structures in place of Boltz-predicted coordinates to assess the impact of poor structure predictions on performance. This model did not outperform Boltz-2-PPI trained using predicted structures, suggesting that structural quality is not the primary performance bottleneck.

For completeness, we also tested the pre-trained Boltz-2 model by representing the peptide as a small molecule (SMILES) -- despite the peptide being an out-of-distribution input for Boltz-2, and not considering the MHC as part of the ligand -- which showed poor performance.

Taken together, these results suggest that Boltz-2 embeddings do not easily yield signal for affinity prediction in this low-data regime, especially when compared to sequence-based alternatives.

\begin{table}[ht]
\centering
\caption{Performance of Boltz-2-PPI and baselines on the TCR3d test set. Reported values are Pearson correlation ($r$) and Spearman correlation ($\rho$).}
\label{tab:tcr3d-results}
\begin{tabular}{lccc}
\toprule
\textbf{Model} & \textbf{Pearson $r$ ($\uparrow$)} & \textbf{Spearman $\rho$ ($\uparrow$)} \\
\midrule
Boltz-2-PPI & 0.153 & 0.091 \\
Boltz-2-PPI (small affinity module) & 0.144 & 0.111 \\
Boltz-2-PPI (true structures) & 0.159 & 0.111 \\
Sequence baseline (ESM2-650M) & \textbf{0.239} & \textbf{0.193}  \\ 
\bottomrule
\end{tabular}
\end{table}

\subsection{Performance on the PPB-affinity Dataset}

We fine-tuned Boltz-2-PPI on the larger PPB-affinity (filtered) dataset~\citep{PPB_enhanced}. A summary of results is presented in Table~\ref{tab:ppb-results}. The structure-based Boltz-2-PPI model underperforms when compared with sequence-based baselines reported in Alsamkary \textit{et al.}~\citep{PPB_enhanced}. This reinforces the conclusions from experiments with the TCR3d dataset that the Boltz embeddings provide weaker signals for affinity prediction relative to direct sequence representations.

To investigate further, we combine Boltz-2-PPI embeddings with sequence embeddings from literature-baselines. We re-train the Prot-T5 and ESM2 sequence models using using published code with default parameters~\citep{PPB_enhanced} (achieving comparable results to those reported on the test set - Appendix \ref{app:retrain}), and combine the final embeddings with representations extracted from the affinity module of our best Boltz-2-PPI model. For the weaker sequence-based model (ESM2-650M-SC), incorporating Boltz-2-PPI features yields modest improvements. The effect is less pronounced for the stronger sequence model (ProtT5-PAD). These results suggest that structural representations learned from Boltz embeddings contain complementary information, albeit significantly less for high-capacity sequence transformers.

\begin{table}[ht]
\centering
\caption{Performance of Boltz-2-PPI and sequence-based models on the PPB-affinity (filtered) test set. Reported values are Pearson correlation ($r$), Spearman correlation ($\rho$), and the root mean squared error (RMSE). Combined models are computed by us, baseline results are taken from ~\citep{PPB_enhanced}.}
\label{tab:ppb-results}
\begin{tabular}{lccc}
\toprule
\textbf{Model} & \textbf{Pearson $r$ ($\uparrow$)} & \textbf{Spearman $\rho$ ($\uparrow$)} & \textbf{RMSE ($\downarrow$)}\\
\midrule
Boltz-2-PPI (fine-tuned, structure only) & 0.338 & 0.357 & 1.362 \\
Sequence baseline (ProtT5-PAD) & 0.48 & 0.51 & 1.42 \\
Sequence baseline (ESM2-650M-SC) & 0.47 & 0.48 & 1.74 \\
Combined (ESM2-650M-SC + Boltz-2-PPI) & 0.487 & 0.483 & 1.367 \\
Combined (ProtT5-PAD + Boltz-2-PPI) & \textbf{0.496} & \textbf{0.515} & \textbf{1.326} \\
\bottomrule
\end{tabular}
\end{table}

\section{Discussion}

Across both the TCR3d and PPB-affinity (filtered) datasets, sequence-based models consistently outperform Boltz-2-PPI. Adapting the Boltz-2 affinity module for protein–protein interactions yielded limited gains, even when training on true structures, suggesting that Boltz representations, while strong for structure prediction, lack the expressiveness needed for affinity regression. In contrast, pre-trained protein language models capture these signals more easily.

Our combined models show that Boltz embeddings can add complementary value to weaker sequence models, pointing to the promise of integrating structural and sequence-derived representations. Progress will likely depend on more sophisticated fusion strategies and larger, more homogeneous affinity datasets.

Finally, while Boltz-2 leverages diverse, multi-fidelity supervision (e.g. docking decoys, quality scores, affinity proxies), we restricted training to curated affinity datasets for controlled benchmarking. Incorporating broader, lower-fidelity binding data could provide richer supervision, paralleling Boltz-2’s strategy and potentially improving generalisation.

\bibliographystyle{unsrt}
\bibliography{neurips_mlsb}

\break
\appendix
\section{Dataset splitting}
\label{app:dataset-split}
\begin{algorithm}[ht]
\caption{ }
\begin{algorithmic}[1]
\STATE \textbf{Input:} Complexes $\mathcal{C} = \{A_1, A_2, \ldots, A_{251}\}$, threshold $\tau = 20.0$, test ratio $r = 0.40$
\STATE \textbf{Output:} Train set $\mathcal{T}$, test set $\mathcal{V}$

\FOR{each pair $(A_i, A_j) \in \mathcal{C} \times \mathcal{C}$ where $i \neq j$}
    \STATE Compute $D(A_i, A_j) = \frac{1}{N_{A_i}} \sum_{k=1}^{N_{A_i}} \min_{\ell} \text{Levenshtein}(S_{A_i,k}, S_{A_j,\ell})$
    \IF{$D(A_i, A_j) \leq \tau$}
        \STATE Add edge $(A_i, A_j)$ to graph $G$
    \ENDIF
\ENDFOR

\STATE Find connected components $\{C_1, C_2, \ldots, C_m\}$ in $G$

\FOR{each remaining component $C_k$}
    \IF{$|\mathcal{V}| < r|\mathcal{C}|$ \textbf{and} $|\mathcal{V}| + |C_k| \leq 1.2 r|\mathcal{C}|$}
        \STATE $\mathcal{V} \leftarrow \mathcal{V} \cup C_k$
    \ELSE
        \STATE $\mathcal{T} \leftarrow \mathcal{T} \cup C_k$
    \ENDIF
\ENDFOR

\end{algorithmic}
\end{algorithm}

\section{Re-trained sequence-based model PPB-affinity performance}
\label{app:retrain}

\begin{table}[ht]
\centering
\caption{Performance re-trained sequence-based models on the PPB-affinity (filtered) test set versus reported values.}
\begin{tabular}{lccc}
\toprule
\textbf{Model} & \textbf{Pearson $r$ ($\uparrow$)} & \textbf{Spearman $\rho$ ($\uparrow$)} & \textbf{RMSE ($\downarrow$)}\\
\midrule
Re-trained (ProtT5-PAD) & 0.484 & 0.506 & 6.055 \\
Re-trained (ESM2-650M-SC) & 0.462 & 0.456 & 1.360 \\
Reported (ProtT5-PAD) & 0.48 & 0.51 & 1.42 \\
Reported (ESM2-650M-SC) & 0.47 & 0.48 & 1.74 \\
\bottomrule
\end{tabular}
\end{table}

\end{document}